\documentclass[conference]{IEEEtran}
\IEEEoverridecommandlockouts

\usepackage{cite}
\usepackage{amsmath,amssymb}
\usepackage{graphics} 
\usepackage{graphics} 
\usepackage{epsfig} 
\usepackage{mathptmx} 
\usepackage{times} 
\usepackage{amsmath} 
\usepackage{amssymb}  
\usepackage{amsfonts,bm}
\usepackage{stfloats}
\usepackage{algorithm}
\usepackage{algorithmic}
\usepackage{tikz}
\usepackage{subcaption}
\usepackage[hyphens]{url}
\usepackage[hidelinks]{hyperref}
\usetikzlibrary{shapes,arrows}
\usepackage{color}
\RequirePackage{filecontents}
\title{{\fontsize{18}{18}\selectfont Safe, Fluent and Acceptable Motion Generation and Execution for Human--Robot Interaction in Manufacturing Environments}}

\author{
Thibaut Lopez$^{1}$, Olivier Aycard$^{1}$, Pierre-Brice Wieber$^{2}$, Mohamed Boua$^{3}$, Christine Jeoffrion$^{3}$%
\thanks{This work is supported by the French National Research Agency in the framework of the ``Investissements d'avenir'' program (ANR-15-IDEX-02).}%
\thanks{$^{1}$ Thibaut Lopez and Olivier Aycard are with GIPSA Lab and Grenoble Institute of Technology, 38000 Grenoble, France {\tt\small thibaut.lopez1@univ-grenoble-alpes.fr, olivier.aycard@grenoble-inp.fr}}%
\thanks{$^{2}$ Pierre-Brice Wieber is with Inria, 38000 Grenoble, France {\tt\small pierre-brice.wieber@inria.fr}}%
\thanks{$^{3}$ Mohamed Boua and Christine Jeoffrion are with LIP/PC2S and Univ. Grenoble Alpes, Univ. Savoie Mont Blanc, 38000 Grenoble, France {\tt\small mohamed.boua@univ-grenoble-alpes.fr, christine.jeoffrion@univ-savoie.fr}}%
}

\IEEEpubid{\raisebox{-11mm}{\begin{minipage}{\dimexpr\textwidth-2\columnsep\relax}\centering
\large This work has been submitted to IEEE for possible publication. Copyright may be transferred without notice, after which this version may no longer be accessible.
\end{minipage}}}

\begin{document}

\maketitle
\IEEEpubidadjcol

\begin{abstract}
Robots operating in human environments must not only ensure physical safety but also exhibit behaviors that are understandable, fluent, and acceptable to human partners. This paper investigates motion generation strategies that combine safety guarantees with interaction quality considerations, such as motion smoothness and human comfort. While the design of robots capable of ensuring safety in shared human-robot environments has enabled closer and more advanced forms of interaction, these new proximity-based tasks require moving beyond purely technical considerations. In particular, robot behavior must also be addressed from psycho-cognitive and social perspectives. In this context, we argue for the relevance of integrating social-aware motion control into robotic systems. First, we identify the motion parameters that influence human perception and operator experience. Then, we implement a Model Predictive Control (MPC) framework that generates four distinct socially-informed robot behaviors. Finally, we conduct a user study to evaluate and validate these behaviors and assess their social impact on non-expert participants. The results demonstrate that variations in robot behavior significantly affect the perceived social acceptability of the system. These findings highlight the importance of incorporating human-centered considerations into motion generation strategies for robots operating in shared environments.
\end{abstract}


\section{Introduction}

Robotic pick-and-place in human-shared environments, that is, in spaces shared with untrained humans and evolving rules defined by users, is a key enabler for service and collaborative robotics. Such capabilities are expected to have a strong impact in domains such as healthcare, logistics, and public spaces. However, the safe deployment of robots operating in close proximity to untrained humans remains a critical bottleneck.

Thanks to advances in mechanical design and control, modern manipulator robots can now safely share their workspace with humans. In particular, these systems are designed such that potential collisions result in a low (5\%) probability of serious injury, even at speeds as high as $2\,\mathrm{m\,s^{-1}}$~\cite{Albu07}. Nevertheless, collisions should still be avoided whenever possible, not only because of the remaining risk, however small, but also because each collision disrupts the tasks of both the robot and the human.

In previous work, we combined human motion perception and short-term prediction~\cite{Zheng22} with control strategies to generate real-time robot trajectories that ensure safety while preserving task continuity and motion fluency~\cite{22:zheng2020online}. With respect to the ISO Technical Specification 15066 on collaborative robots, and unlike existing solutions based on emergency stops or fixed safety sensors~\cite{Li24}, our approach combines aspects of \emph{speed and separation monitoring} when actively avoiding collisions, with aspects of \emph{power and force limiting} if a collision eventually occurs. The focus of our approach is on active collision avoidance, while we rely on the manipulator robot's mechanical and control design to mitigate the risk of injury in the event of contact. Alternative approaches address this latter aspect more directly, for example by explicitly controlling the forces involved during collision~\cite{Svarny2019IROS}.

While such approaches significantly improve physical safety, safety alone is not sufficient for effective human--robot collaboration. In real-world shared environments, robots must not only avoid collisions but also interact with humans in ways that are understandable, predictable, and comfortable. Recent research in human--robot interaction therefore emphasizes the importance of \emph{interaction fluency}, defined as the ability of humans and robots to coordinate their actions smoothly and continuously during a shared task \cite{Hoffman2019}. Motion characteristics such as smoothness, anticipation, and consistency with human expectations are essential for enabling humans to understand robot intentions and adapt their behavior accordingly.

Beyond fluency, \emph{acceptability} has emerged as a key factor in the deployment of collaborative robots. The perceived comfort, trust, and social perception of robot behavior strongly influence the willingness of users to interact with robotic systems \cite{Lasota17}. Parameters such as robot speed, interpersonal distance, and trajectory smoothness can significantly impact human perception of robot behavior and influence the quality of interaction.

These observations suggest that the design of robot motion in human-shared environments should jointly address three complementary objectives: ensuring physical safety, maintaining task efficiency, and generating motion that is both fluent and socially acceptable. In this paper, we build upon predictive safety-aware motion generation approaches and extend them toward a human-aware framework that explicitly incorporates interaction fluency and acceptability constraints.

The contributions of this paper are threefold:

\begin{itemize}
\item First, we identify key motion parameters that influence human perception during human--robot interaction and derive a set of representative interaction behaviors associated with different social attitudes;

\item Second, we design a motion control strategy capable of implementing these interaction behaviors while preserving the safety guarantees provided by our predictive control framework;

\item Third, we evaluate the impact of these behaviors on the users' perceived workload and social perception in real-world experiments involving non-expert users.
\end{itemize}

The remainder of this paper is organized as follows. First, we analyze the motion parameters that influence human perception and define representative interaction behaviors. Then, we briefly summarize the predictive control framework previously developed to ensure safety, and describe its extension to implement these interaction behaviors. Finally, we present experimental results obtained with non-expert users interacting with the robot in realistic conditions.

\section{Behavior Parameterization}
\label{sec:behavior_parameterization}
In order to parameterize robot motion during human--robot interaction, we consider two fundamental control variables: the \textbf{maximum translational velocity} $v_{max}$ and the \textbf{maximum interaction distance} $d_{max}$. These parameters are selected as they directly modulate both the dynamic properties of robot motion and the proxemic relationship between the robot and the human.

The choice of these parameters is strongly supported by prior work in human--robot interaction. In shared workspaces, the kinematic and geometric properties of robot motion have a direct impact on user comfort, perceived safety, and collaboration efficiency. In particular, robot velocity and interpersonal distance are identified as primary factors influencing human perception and interaction quality \cite{Lasota17,Sisbot2007}.

Several studies have shown that robot velocity directly affects perceived safety and anticipation capabilities. Lower velocities improve motion predictability and allow the human to better anticipate robot behavior, whereas higher velocities tend to increase stress and reduce perceived safety \cite{Lasota17}.

Moreover, the distance parameter influences the degree of overlap between human and robot workspaces, which has been shown to affect cognitive load: larger overlaps require continuous monitoring from the human, increasing mental effort and reducing trust in the system \cite{Howell2019}.

Beyond velocity and distance, motion dynamics such as acceleration profiles also impact perception. Abrupt changes in motion can induce surprise or discomfort, whereas smooth and progressive adaptations improve legibility and facilitate coordination \cite{Dragan2013,Hoffman2019}. These elements highlight that motion is not only functional but also acts as a communication channel conveying robot intent.

In this work, we deliberately focus on velocity and interaction distance as primary parameters, as they are both fundamental, interpretable, and directly controllable within our system. They also constitute key dimensions of motion expressivity and proxemic interaction, making them suitable for systematically generating distinct robot behaviors.

Formally, we define:
\begin{equation}
v_{max} \in \mathbb{R}^+ \quad \text{and} \quad d_{max} \in \mathbb{R}^+
\end{equation}

Using these two parameters, we construct a discrete set of four motion regimes by combining two levels of each parameter (low/high velocity and short/long interaction distance):

\begin{itemize}
    \item \textbf{Low $v_{max}$, Short $d_{max}$}: reactive and cautious behavior with late adaptation.
    \item \textbf{Low $v_{max}$, Long $d_{max}$}: anticipatory and conservative behavior promoting smooth interaction.
    \item \textbf{High $v_{max}$, Short $d_{max}$}: efficiency-driven behavior with reduced legibility.
    \item \textbf{High $v_{max}$, Long $d_{max}$}: anticipatory yet efficient behavior balancing performance and interaction quality.
\end{itemize}

This $2 \times 2$ parameterization enables a controlled exploration of the trade-off between efficiency, safety, cognitive load, and interaction quality. It is grounded in established findings in human--robot interaction, which highlight the central role of velocity, distance, and motion dynamics in shaping human perception of robot behavior. The subsequent user study, in section~\ref{sec:behavior_parameterization} aims to verify if these intended behaviors translate into actual differentiated human perceptions.

Beyond classical evaluation criteria in robotics, such as safety, efficiency, or trajectory smoothness, recent research in human--robot interaction has highlighted that robot motion inherently conveys socio-affective cues that shape human perception. This perspective is formalized by the concept of movement prosody, which draws an analogy between motion dynamics and vocal prosody in human communication. In particular, prior work has shown that robot behavior can be assessed through bipolar perceptual dimensions derived from expressive human movement studies. These dimensions aim to capture how robot motion is interpreted in terms of attitude, intention, and social meaning, beyond purely geometric or kinematic properties.

In line with this perspective, we evaluate robot behavior through $10$ bipolar adjective scales: \emph{aggressive--gentle}, \emph{authoritarian--polite}, \emph{confident--hesitant}, \emph{solid--fragile}, \emph{strong--weak}, \emph{smooth--rough}, \emph{rigid--flexible}, \emph{tender--sensitive}, \emph{inspires confidence--does not inspire confidence} and \emph{sympathetic--antipathetic}. Taken together, these dimensions cover both physical qualities of motion and higher-level social interpretations. Some scales primarily reflect the dynamic and embodied properties of the movement, such as \emph{smooth--rough}, \emph{rigid--flexible}, \emph{solid--fragile}, or \emph{strong--weak}. Others capture interpersonal or affective impressions, such as \emph{authoritarian--polite}, \emph{aggressive--gentle}, \emph{sympathetic--antipathetic}, or \emph{tender--sensitive}. Finally, dimensions such as \emph{confident--hesitant}, \emph{self-assured--doubtful}, and \emph{inspires confidence--does not inspire confidence} characterize the degree to which the robot is perceived as assured in its motion and able to elicit trust from the human partner.

This choice allows us to examine not only whether a given motion strategy is safe and efficient, but also how it is interpreted by human users. However, unlike prior work that mainly seeks to identify or validate perceptual dimensions, our objective is to analyze how controlled motion parameters, namely velocity modulation and interaction distance, influence these perceptions.

\section{Hierarchical Control Architecture for Behavior Generation}
\begin{figure}[!t]
\begin{center}
	\includegraphics[width=0.5\textwidth]{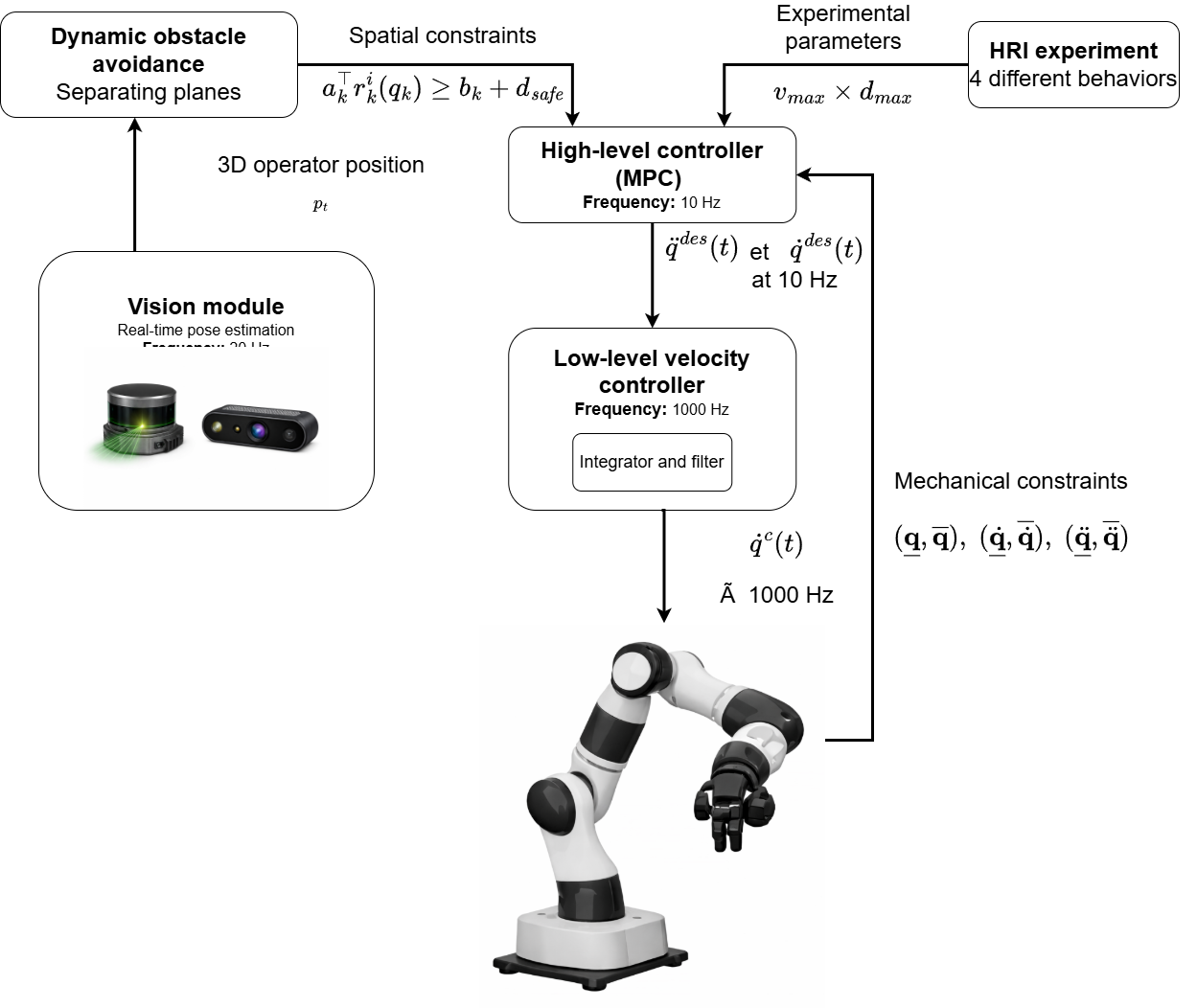}
\end{center}
\caption{Abstract flow diagram of control architecture.}
\label{fig:control_archiitecture}
\end{figure}

The goal of our scheme is to generate a collision-free trajectory for a cobot (e.g., a 7-DoF manipulator cobot) that has to perform a task in a workspace shared with a human worker. This scheme is composed of 3 parts, as shown in Figure~\ref{fig:control_archiitecture}: (i) a collision-free trajectory generation module (section~\ref{sec:high-level}), (ii) a human motion prediction module (section~\ref{sec:detection}) and (iii) a low-level robot motion control module (section~\ref{sec:low-level}).

\section{A collision free trajectory module}
\label{sec:high-level}
In this section, we summarise how to compute a collision-free trajectory based on the perception module and the cobot's dynamics. Also, we emphasise the role of the terminal constraint to guarantee safety. This terminal constraint provides a \textit{passive motion safety} guarantee \cite{26:bouraine2012provably}, which means that if a collision occurs, the cobot is at rest at the time of the collision so that it doesn't inject its own kinetic energy. In the following, we recall the main equations from the MPC approach developed in our previous work \cite{22:zheng2020online}.

\subsection{Separating plane optimisation}
\begin{figure}[!t]
\begin{center}
	\includegraphics[width=0.9\columnwidth]{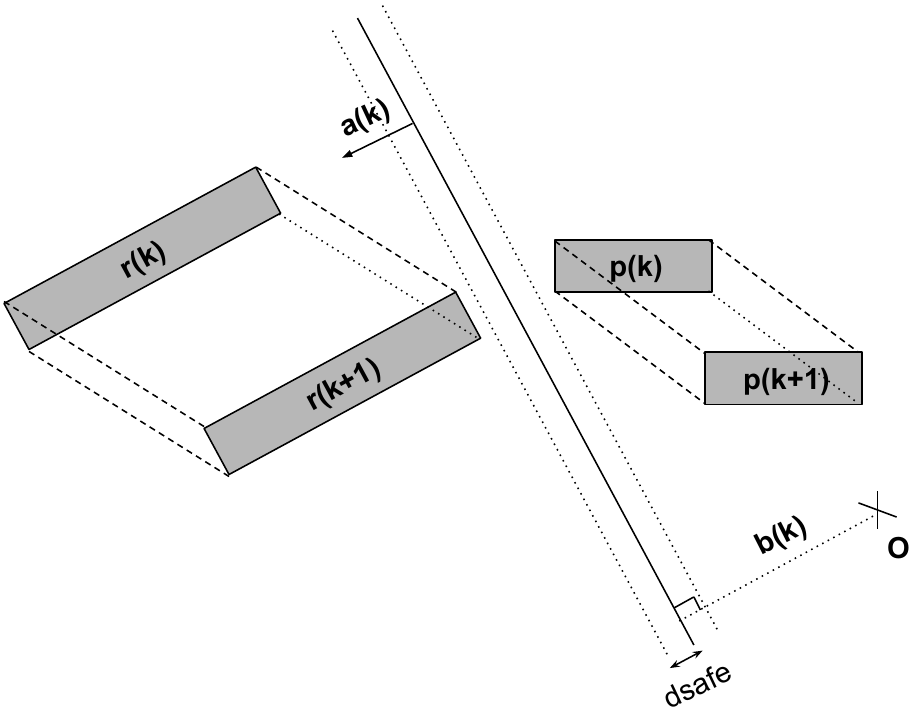}
\end{center}
\caption{An illustration of separating plane between two objects.}
\label{fig:separating_plane}
\end{figure}

As illustrated in Figure~\ref{fig:separating_plane}, if there exists at the prediction time $k \in \mathbb{N}$ a plane defined by a normal vector $a_k \in \mathbb{R}^3$ and a scalar constant $b_{k} \in \mathbb{R}$ such that all vertices $\mathbf{p}_k^{j}$ related to the human stay on one side between instants $k$ and $k+1$ while all vertices $r^{i}$ related to the cobot stay on the other side, then we have evidence that they don't collide over this interval of time. Here, $j\in\{1,\ldots N_p \}$ and $i\in\{1,\ldots N_r\}$ where $N_{p}$ and $N_{r}$ are the number of vertices associated with the human and the cobot. The distance $d \in \mathbb{R}$ is the geometric margin optimized during the separating-plane computation: it measures how far the robot vertices lie from the plane once the human vertices are constrained to remain on the other side. In the following, $a_k^T$ denotes the transpose of the normal vector $a_k$.

\begin{subequations}\label{eq:lp}
\begin{align}
\mathop{\mbox{min.}}_{a_k,\,b_k,\,d}\ 
&-d+\alpha d^2+\beta\|a_k-a_k^p\|^2+\beta||b_k-b_k^p||^2 \label{eq:costlp} \\  
\mbox{s.t.}\ \ 
&\forall\,j\in\{1,\ldots N_p \},\ a_k^T\mathbf{p}_k^j\leq b_k, \label{eq:pk_1} \\ 
&\forall\,j\in\{1,\ldots N_p \},\ a_k^T\mathbf{p}_{k+1}^j\leq b_k, \label{eq:pk_2} \\ 
&\forall\,i\in\{1,\ldots N_r \},\ a_k^Tr_k^i\geq b_k+d, \label{eq:rk_1} \\ 
&\forall\,i\in\{1,\ldots N_r \},\ a_k^Tr_{k+1}^i\geq b_k+d, \label{eq:rk_2} \\ 
&-1_3\leq a_k\leq 1_3, \label{eq:norm_1}\\
&1-\varepsilon\leq a_k^T\,a_k^p\leq1 \label{eq:norm_2}
\end{align}
\end{subequations}
We want to maximise the distance $d$ between the separating plane and the cobot. Given the formulation as a minimisation problem, we include the term $-d$ in the cost function, Equation~\ref{eq:costlp}. This quantity $d$ is an intermediate result of the plane-search problem and should be distinguished from the fixed safety margin used later in the MPC constraints. The following term in the cost function smooths the variations of separating planes, with $a_{k}^{p}$ and $b_{k}^{p}$ the separating plane parameters obtained at the previous sampling time, and some small weights $\alpha$ and $\beta$. Finally, we use constraints, Equations~(\ref{eq:norm_1}-\ref{eq:norm_2}), to approximate a nonlinear constraint to bound the vector $a_{k}$ to a unit norm. Where $1_3 \in \mathbb{R}^{3}$ is a row vector of ones.

\vspace{-0.5em}
\subsection{Optimal motion generation}

Once we have a sequence of separating planes parameters, we can include them in our MPC scheme to compute an optimal collision-free trajectory:
\begin{subequations}\label{eq:qp}
\begin{align}
\mathop{\mbox{min.}}_{\textbf{u}}\ \ 
&\sum_{k=0}^{N-1}\|s_{k+1}-s_{k+1}^\mathit{des}\|^2_{Q}+\|u_k-u_k^\mathit{des}\|^2_{R} \label{eq:cost_QP}\\  
\mbox{s.t.}\ \ 
&\forall\,k\in\{0,\ldots N-1\},\ \underline{u}\leq u_k\leq\overline{u},\\
&\forall\,k\in\{1,\ldots N\},\ \underline{q}\leq q_k\leq\overline{q},\\
&\forall\,k\in\{1,\ldots N-1\},\ \underline{\dot{q}}\leq\dot{q}_k\leq\overline{\dot{q}},\\
&\dot{q}_N=0, \label{eq:qp_terminal} \\ 
&\forall\,k\in\{0,\ldots N-1\},\ \forall\,i,\nonumber\\
&\ \ \ \ \ a_k^Tr_k^i(q_k^p)+a_k^TJ(q_k^p)(q_k-q_k^p)\geq b_k+d_\mathit{safe}, \label{eq:avoidance_1}\\
&\forall\,k\in\{0,\ldots N-1\},\ \forall\,i,\nonumber\\
& a_k^Tr_k^i(q_{k+1}^p)+a_k^TJ(q_{k+1}^p)(q_{k+1}-q_{k+1}^p)\geq b_k+d_\mathit{safe} \label{eq:avoidance_2}
\end{align}
\end{subequations}
Where $q_k \in \mathbb{R}^n$ and $\dot{q}_k \in \mathbb{R}^{n}$ are respectively the joint position and velocity, with $n$ the number of degrees of freedom. The state $s_{k} \in \mathbb{R}^{2n}$ includes $q_k$ and $\dot{q}_k$, and the reference state $s_k^\mathit{des} \in \mathbb{R}^{2n}$ is defined similarly. The control input $u_{k} \in \mathbb{R}^{n}$ is the joint acceleration of the cobot, while $u_k^\mathit{des}$ denotes its reference value.
The safety distance $d_\mathit{safe}$ represents a prescribed buffer zone between the robot and the predicted human position. Unlike $d$, which is optimized when computing the separating plane, $d_\mathit{safe}$ is a design parameter imposed in the MPC constraints to enforce a minimum clearance during execution.
\begin{algorithm}[t]
    \caption{CollisionFreeTrajectoryComputation} 
    \hspace*{0.02in} {Input:} 
    $U_k^{p}$, $S_k$, $a_k^{p}$, $b_k^{p}$\\
    \hspace*{0.02in} {Output:} 
    $U_k$
    \begin{algorithmic}[1]
    \STATE $i \gets 0$
    \WHILE{$\|U_k - U_k^{p}\|^2$ or $i \leq k$}
    \STATE $U_k^{p} \gets U_k$
    \STATE Update robot parameters
    \STATE $\{a,b\} \gets$ solve Eq.~(\ref{eq:lp}) for $k \in \{0,\ldots,N-1\}$
    \STATE $\{U_k\} \gets$ solve Eq.~(\ref{eq:qp})
    \STATE $i \gets i + 1$
    \ENDWHILE
       
    \end{algorithmic}
\end{algorithm}



      
Our prediction horizon has a length $N \in \mathbb{N}$. The cost function, Equation~\ref{eq:cost_QP}, is designed to track a reference state trajectory $s_{k}^{\mathit{des}}$ together with a reference acceleration input $u_{k}^{\mathit{des}}$, while $\underline{q}$, $\overline{q}$, $\underline{\dot{q}}$, $\overline{\dot{q}}$, $\underline{u}$, $\overline{u}$ indicate minimum and maximum joint positions, velocities, and accelerations (we assume that $\underline{\dot{q}}\leq0\leq\overline{\dot{q}}$ and $\underline{u}\leq0\leq\overline{u}$). The terminal constraint, Equation~\ref{eq:qp_terminal}, ensures that the cobot is at rest at the end of the prediction horizon in order to provide a passive motion safety guarantee, making sure that the cobot is able to stop and stay at rest before any collision happens in the future. Equations~\ref{eq:avoidance_1}-\ref{eq:avoidance_2} introduce the collision avoidance constraints based on separating planes, which are computed by linearising the kinematics of the cobot around the previously computed trajectory:
\begin{equation}
r_{k}^{i} = r^{i}(q_k) \approx r^{i}(q_k^{p}) + J(q_{k}^{p})(q_k - q_k^{p}) 
\end{equation} 

\subsection{Behavior Encoding}

Interaction behaviors are encoded through two parameters $v_{max}$ and $d_{max}$. These parameters directly affect the velocity bounds $\overline{\dot{q}} \propto v_{max}$ and the safety distance $d_\mathit{safe} \propto d_{max}$.

The safety distance is directly proportional to the interaction distance parameter $d_{max}$ defined in Section~\ref{sec:behavior_parameterization}. Larger values of $d_{max}$ result in more conservative (anticipatory) motion, with increased clearance from human predictions. This safety margin is used in the collision-avoidance constraints (Equations~\ref{eq:avoidance_1}-\ref{eq:avoidance_2}) to ensure that the robot maintains a minimum separation from the human trajectory.

By selecting different combinations of $(v_{max}, d_{max})$, the controller generates four distinct behaviors, modulating anticipation, reactivity, and efficiency. 

\section{Human Pose Detection and Human Hand Motion Prediction}
\label{sec:detection}
The collision-avoidance scheme described in Section~\ref{sec:high-level} relies fundamentally on accurate predictions of future human positions to compute separating planes and enforce collision-avoidance constraints in the MPC formulation. Specifically, the predicted human hand positions feed into the separating plane optimization (Equations~\ref{eq:pk_1}-\ref{eq:pk_2}) and ultimately shape the safety constraints (Equations~\ref{eq:avoidance_1}-\ref{eq:avoidance_2}) that govern the robot's trajectory. The quality and accuracy of these predictions thus directly determine the effectiveness of the motion control strategy described above. In this section, we summarize our work on perception~\cite{Zheng22} and describe how we estimate the 3D position of the human hand from RGB-D data and use the resulting trajectory as input to a prediction model.

The collision-avoidance scheme requires a short-horizon prediction of future human positions. We therefore estimate the 3D position of the human hand from RGB-D data and use the resulting trajectory as input to a prediction model.

\subsection{Human Pose Detection}
\begin{figure}[!t]
\begin{center}
	\includegraphics[width=0.95\columnwidth]{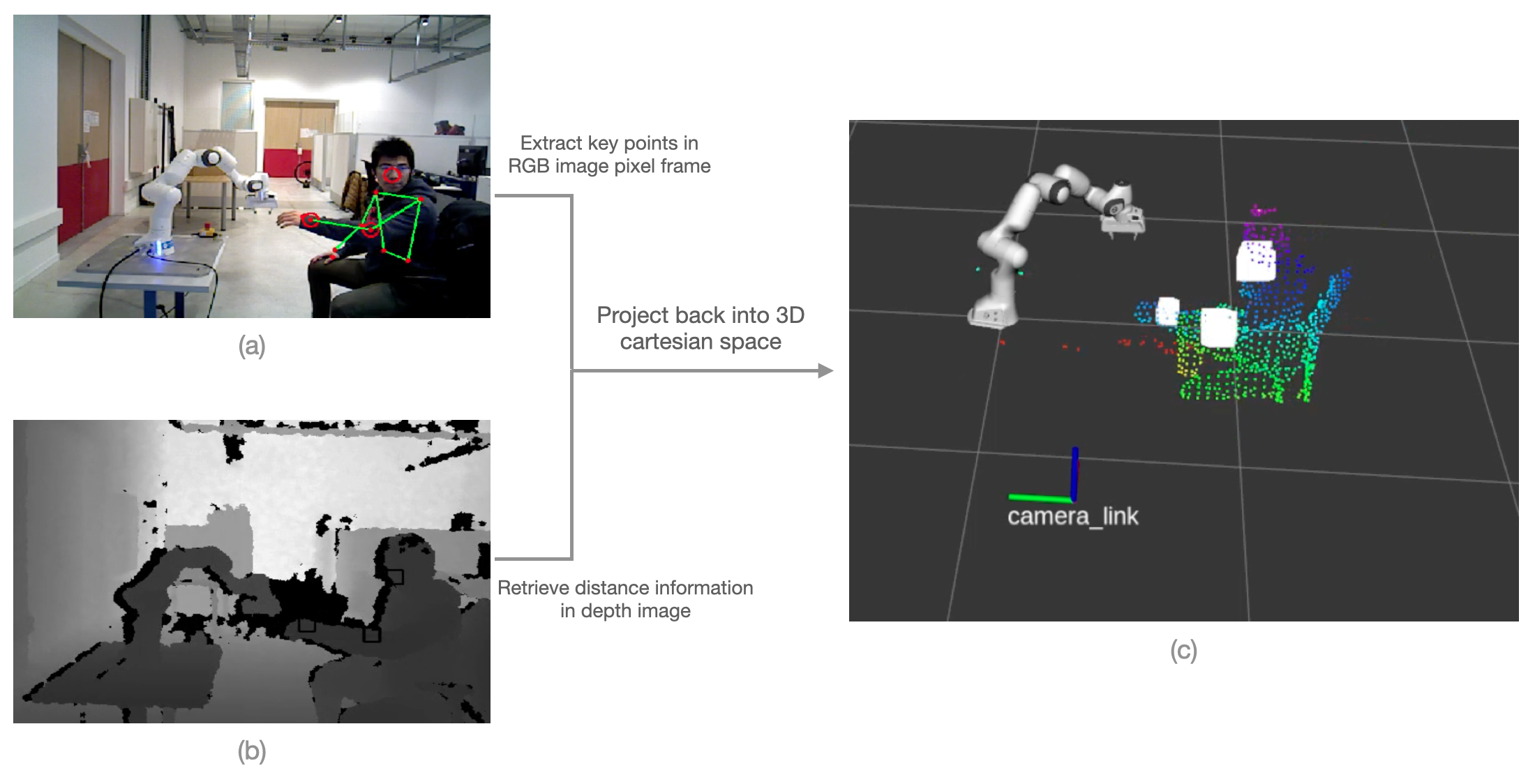}
\end{center}
\caption{Example demonstration of RGB-D image for mapping human hand in 3D space, a) shows RGB image on which hand joins are detected, b) shows the bounding box around the points in the depth image, and c) shows the mapping of points in 3D space.}
\label{fig:human_detection}
\end{figure}

The perception system relies on a Realsense L515 camera providing synchronized colour and depth images. Human keypoints are first detected in the RGB image with lightweight pose-estimation methods such as OpenPose~\cite{38:cao2019openpose} or MediaPipe~\cite{39:lugaresi2019mediapipe}; examples are shown in Figure~\ref{fig:human_detection}. The corresponding depth values are then associated to the detected pixels after camera calibration~\cite{40:nowak2021point}, and the 3D Cartesian position is recovered with the pinhole projection model.


\subsection{Human Hand Motion Prediction}
\label{sec:prediction}
From the measured 3D hand positions, we predict a short sequence of future positions. To stay consistent with Section~\ref{sec:high-level}, we denote by $p_t \in \mathbb{R}^3$ the 3D position of the human hand at time $t$. The hand dynamics are modeled in discrete time as
\begin{subequations}
\begin{align}
p_{t+1} &= g(p_t, w_t) \label{eq:human_dynamic1}\\
z_t &= h(p_t) \label{eq:human_dynamic2}
\end{align}
\end{subequations}
where $w_t$ represents unmeasured inputs and $z_t$ is the observed hand position. Rather than predicting only one step ahead, we seek a mapping from a short observation window $\mathbf{p}=\{p_t,\dots,p_{t-L}\}$ to a future sequence $\hat{\mathbf{p}}=\{\hat{p}_{t+1},\dots,\hat{p}_{t+T}\}$.

Because human motion is nonlinear and affected by unmeasured factors, the predictor is implemented as an encoder-decoder LSTM network. Past observations are encoded through stacked recurrent layers, and the decoder outputs a multi-step prediction $\hat{\mathbf{p}}$ of the future hand trajectory. This predicted trajectory is then passed to the motion planner to enforce collision-avoidance constraints in real time.
In the end, the predicted positions $\hat{p}_{t+\ell}$ are provided to the separating plane computation and instantiate the human vertices $\mathbf{p}_k^j$ used in Equations~\ref{eq:pk_1} and \ref{eq:pk_2}.

\section{Low-Level Control: Velocity Tracking and Execution}
\label{sec:low-level}
The low-level controller is responsible for executing in real time the trajectory computed by the MPC. While the high-level predictive controller operates on the horizon and enforces collision-avoidance and terminal safety constraints, the low-level loop runs at a higher frequency (typically $100$--$1000\,\mathrm{Hz}$) and converts the planned joint motion into actuator commands. Its role is therefore not to re-plan the motion, but to ensure that the robot accurately tracks the admissible trajectory generated by the upper layer despite discretization, actuation delays, and small execution errors.

In order to keep the notation consistent with Section~\ref{sec:high-level}, we denote by $q^\mathit{des}(t)$, $\dot{q}^\mathit{des}(t)$, and $\ddot{q}^\mathit{des}(t)$ the position, velocity, and acceleration references provided by the high-level controller after interpolation of the MPC solution at a period $\Delta t$. The measured joint position and velocity are denoted by $q(t)$ and $\dot{q}(t)$. The low-level controller then computes a command $\dot{q}^c(t)$ to be sent to the robot servo loop at a period $\delta t$.

\subsection{Velocity Integration and Command Generation}

The MPC described in Section~\ref{sec:high-level} provides a sequence of joint accelerations $u_k$ together with the associated state trajectory $(q_k,\dot{q}_k)$. At execution time, these discrete references are converted into continuous-time commands over one control period. 
In the velocity-controlled implementation, the commanded velocity must follow the state evolution predicted by the MPC, in particular the desired velocity $\dot{q}^{des}$. To ensure consistency between the predictive model and the low-level controller, the desired acceleration $\ddot{q}^{des}$ is integrated over the current control interval using $\dot{q}^{des}$ as the initial condition.

In practice, $\ddot{q}^{des}$ corresponds to the acceleration command optimized by the MPC, i.e., $\ddot{q}^{des}(t) \approx u_k$ over the current sampling interval. Since the MPC update rate is not strictly periodic and may deviate from the nominal period $\Delta t$, the actual integration time step $\delta t$ may differ from $\Delta t$. Using $\dot{q}^{des}$ as the initial condition ensures that the commanded velocity remains consistent with the linearized dynamics used in the MPC prediction model.

This results in the following control law:
\begin{equation}
\label{equation_vitesse_integration_bas_niveau}
\dot{q}^c(t) = \dot{q}^{des}(t) + \ddot{q}^{des}(t)\,\delta t.
\end{equation}

The value computed by \eqref{equation_vitesse_integration_bas_niveau} is then fed to the robot servo-loop. Is integrated in this system a PID controller to guarantee that the robot follows the given joint speed $\dot{q}^c(t)$

Overall, the low-level controller ensures faithful realization of the motion behaviors defined at the high level. The interaction style is determined upstream through the parameters $v_{max}$ and $d_{max}$ and through the MPC constraints, whereas the low-level loop guarantees that these planned behaviors are executed smoothly and accurately on the real robot. This separation between predictive planning and fast feedback control is important: it preserves the safety and interaction properties established at the planning level while providing the robustness required for physical execution.









\section{Experiment 1: Preliminary validation of the four behaviors}
We first ran a preliminary robotic experiment to verify that the controller could generate the four behaviors introduced in Section~\ref{sec:behavior_parameterization}. In a fixed pick-and-place task, only $v_{max}$ and $d_{max}$ were varied, while the rest of the control architecture remained unchanged.

The four tested conditions are listed in Table~\ref{tab:behaviors}. Overall, the controller produced clearly differentiated motions while preserving task execution and safety. Videos of the experiment are available in this \href{https://youtube.com/playlist?list=PL8ZyzBKlMS520Njc_XROIVTgbS9IDw_n_}{YouTube playlist}.

\begin{table}[t]
\caption{Behavioral conditions used in the experiments}
\label{tab:behaviors}
\centering
\scriptsize
\setlength{\tabcolsep}{2pt}
\begin{tabular}{|c|p{2.2cm}|c|p{1.6cm}|}
\hline
Condition & Behavior name & Velocity profile & Interaction distance \\
\hline
B1 & Reactive and cautious & Low & Short / near \\
\hline
B2 & Anticipatory and conservative & Low & Long / far \\
\hline
B3 & Efficiency-driven & High & Short / near \\
\hline
B4 & Anticipatory yet efficient & High & Long / far \\
\hline
\end{tabular}
\end{table}

\begin{figure*}[!t]
\centering
\includegraphics[width=0.9\linewidth]{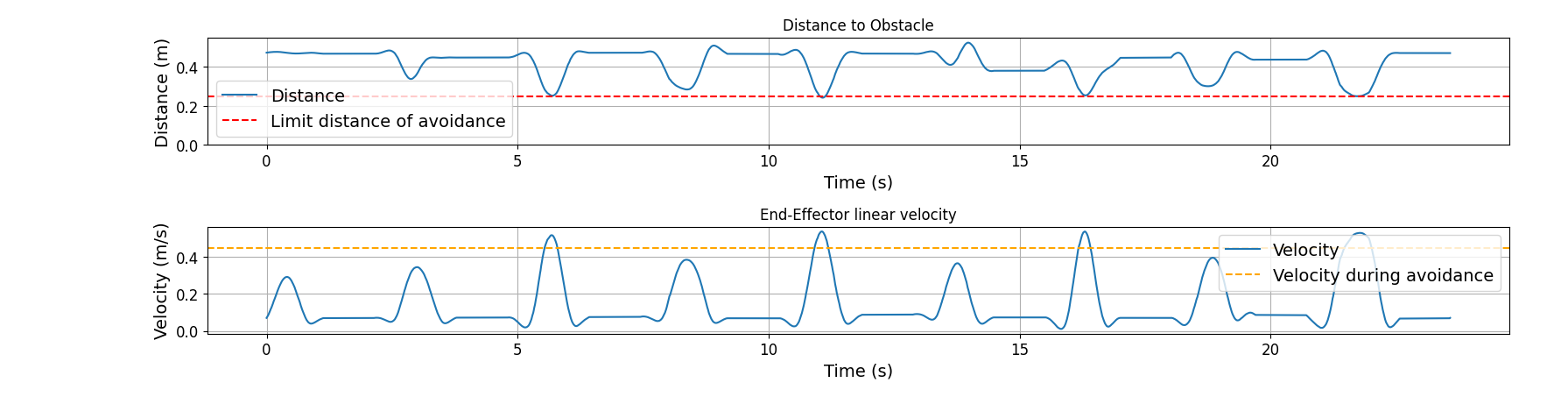}
\caption{The robot's behavior when efficiency-driven.}
\label{fig:behavior_efficiency}
\end{figure*}

\begin{figure}[!t]
\centering
\includegraphics[width=0.7\columnwidth]{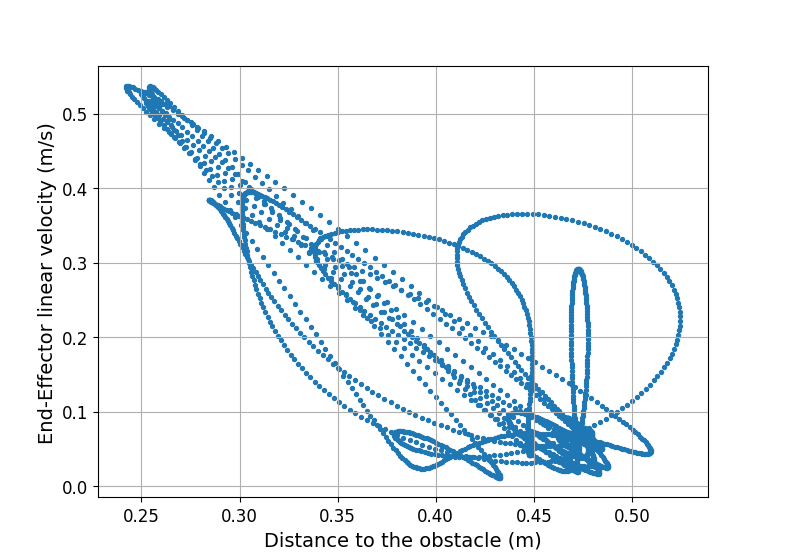}
\caption{Velocity profile during obstacle avoidance showing acceleration as the robot approaches the obstacle.}
\label{fig:velocity_distance}
\end{figure}

An example of the efficiency-driven behavior is shown below. When the human–robot distance drops below the avoidance threshold, the robot accelerates to perform the evasive maneuver, as shown in Fig.~\ref{fig:behavior_efficiency}. This behavior is consistent with the velocity profile depicted in Fig.~\ref{fig:velocity_distance}, where the robot velocity increases as the distance to the human decreases. This result highlights the efficiency-driven nature of the proposed control strategy, which favors rapid avoidance when the interaction distance becomes critical.

Qualitatively, these preliminary trials showed that the controller can be used as a behavior generator and not only as a safety filter. Larger values of $d_{max}$ produced earlier and more conservative deviations, while the velocity profile mainly affected the temporal appearance of the motion, with accelerating behaviors appearing smoother and more reactive. This first experiment therefore validates the practical relevance of the two-dimensional parameterization and provides the basis for the user study reported in the next section.

\section{Experiment 2: Human--robot interaction study}
\label{sec:experiment2}
The second experiment assessed the acceptability of the generated behaviors from a human-centered perspective\footnote{The experiment was submitted to the Grenoble Alpes Research Ethics Committee (CERGA) for consultation: \href{https://www.univ-grenoble-alpes.fr/universite/engagements/ethique-et-deontologie/comite-d-ethique-pour-la-recherche-grenoble-alpes/}{website}.}. More specifically, it examined whether velocity modulation and interaction distance affect i) subjective workload and ii) the social perception of the robot. This distinction matters because a motion strategy may be geometrically safe while still being uncomfortable or difficult to interpret~\cite{Hoffman2019,Lasota17}.

\subsection{Protocol}

The study involved $111$ participants in a repeated-measures $2 \times 2$ design. All participants provided informed consent, and data were stored anonymously. Each participant evaluated the four conditions resulting from the combination of the two factors: acceleration versus deceleration and short versus long interaction distance. The sample included $78$ women ($70.27\%$) and $33$ men ($29.73\%$), aged $18$ to $68$ years ($M = 25.16$, $SD = 9.82$), yielding $444$ observations in total. The sample included $78$ women ($70.27\%$) and $33$ men ($29.73\%$), with ages ranging from $18$ to $68$ years ($M = 25.16$, $SD = 9.82$). 

\begin{table}[t]
\caption{Summary of the user-study protocol}
\label{tab:protocol}
\centering
\begin{tabular}{|l|l|}
\hline
Participants & $111$ \\
\hline
Design & Repeated-measures $2 \times 2$ \\
\hline
Factors & Velocity $\times$ interaction distance \\
\hline
Velocity levels & Deceleration / acceleration \\
\hline
Distance levels & Short / long \\
\hline
Conditions per participant & $4$ \\
\hline
Total observations & $444$ \\
\hline
Workload measure & NASA-TLX (global and weighted) \\
\hline
Perception measure & $10$ bipolar adjective scales \\
\hline
\end{tabular}
\end{table}

The experimental logic was as follows. After interacting with the robot under each condition, participants were asked to evaluate both the experienced workload and their perception of the robot. Workload was assessed with the NASA-TLX questionnaire. In addition to the usual global score, a weighted NASA-TLX score was also computed, so as to account for the relative importance assigned by each participant to the six workload dimensions. This part of the protocol was intended to capture whether one type of robot behavior required more effort, more monitoring, or generated more stress than another.

Perception of the robot was assessed through $10$ bipolar dimensions on a four-point scale. The evaluated dimensions were: \emph{aggressive--gentle}, \emph{authoritarian--polite}, 
\emph{confident--hesitant}, \emph{solid--fragile}, \emph{strong--weak}, \emph{smooth--rough}, \emph{rigid--flexible}, \emph{tender--sensitive}, 
\emph{inspires confidence--does not inspire confidence} and \emph{sympathetic--antipathetic}. These dimensions were selected to characterize not only physical comfort, but also the social interpretation of the robot motion. The associated analyses were conducted with linear mixed models in order to account for the repeated-measures structure of the data.

\subsection{Measured variables}

The two families of dependent variables correspond to two different hypotheses. The first hypothesis is related to \emph{interaction cost}: if one of the behaviors is too abrupt, too close, or too difficult to anticipate, it should increase workload scores. The second hypothesis is related to \emph{interaction meaning}: even if all behaviors remain similarly easy to handle, they may nevertheless convey different impressions in terms of confidence, smoothness, robustness, politeness, or sympathy. This distinction is central to the contribution of the paper, because it separates what the user experiences functionally from how the user interprets the motion socially, in line with prior HRI studies showing that kinematic cues affect both workload-related coordination and social interpretation~\cite{Howell2019,Dragan2013}.

\subsection{Results on workload}

At the descriptive level, workload scores remained relatively low across all conditions. The mean global NASA-TLX score over the whole study was $2.47$, and the mean weighted NASA-TLX score was $2.54$, indicating a globally low to moderate workload. Table~\ref{tab:nasa_desc} reports the descriptive statistics by condition. In both versions of the score, the near-distance conditions yielded slightly higher averages than the far-distance conditions, and decelerating conditions yielded slightly higher averages than accelerating ones. However, these differences remained small.

\begin{table}[t]
\caption{NASA-TLX scores by experimental condition}
\label{tab:nasa_desc}
\centering
\begin{tabular}{|l|c|c|}
\hline
Condition & Global NASA-TLX & Weighted NASA-TLX \\
\hline
Deceleration + near & $2.56 \; (1.09)$ & $2.65 \; (1.20)$ \\
\hline
Deceleration + far & $2.44 \; (1.07)$ & $2.50 \; (1.14)$ \\
\hline
Acceleration + near & $2.50 \; (1.12)$ & $2.59 \; (1.22)$ \\
\hline
Acceleration + far & $2.38 \; (1.06)$ & $2.43 \; (1.16)$ \\
\hline
\end{tabular}
\end{table}

The inferential analysis confirmed that workload was not significantly affected by the manipulated factors. For the global NASA-TLX score, neither velocity ($p=.353$), nor interaction distance ($p=.491$), nor their interaction ($p=.340$) reached significance. The same conclusion held for the weighted score, with non-significant effects of velocity ($p=.457$), distance ($p=.232$), and interaction ($p=.486$). In practical terms, this means that none of the four behaviors made the interaction substantially more demanding for the participants. The control parameterization therefore appears able to modify the style of the robot motion without degrading the subjective ease of the interaction, which is consistent with the idea that variations in spatial overlap and motion coordination do not necessarily translate into a large cognitive cost when the interaction remains predictable and well structured~\cite{Howell2019,Hoffman2019}.

\subsection{Results on robot perception}

The results were more contrasted for the perceptual variables. While workload remained globally stable, several dimensions of robot perception were sensitive to the velocity profile. Table~\ref{tab:perception_desc} reports descriptive means for a subset of the most informative perceptual dimensions. A clear pattern emerges: accelerating conditions tend to receive more favorable ratings on dimensions related to confidence, solidity, strength, smoothness, and assurance.

\begin{table}[t]
\caption{Selected perceptual ratings by condition (mean (SD))}
\label{tab:perception_desc}
\centering
\scriptsize
\setlength{\tabcolsep}{2pt}
\begin{tabular}{|p{2.25cm}|c|c|c|c|}
\hline
Dimension & D+N & D+F & A+N & A+F \\
\hline
Conf.--hes. & $2.55$ ($0.98$) & $2.52$ ($0.97$) & $2.25$ ($0.96$) & $2.23$ ($0.99$) \\
\hline
Solid--frag. & $2.14$ ($0.85$) & $2.08$ ($0.94$) & $1.94$ ($0.82$) & $2.05$ ($0.87$) \\
\hline
Strong--weak & $2.03$ ($0.78$) & $2.13$ ($0.79$) & $1.94$ ($0.74$) & $1.94$ ($0.81$) \\
\hline
Smooth--rough & $2.39$ ($0.91$) & $2.39$ ($0.91$) & $2.31$ ($0.89$) & $2.26$ ($0.83$) \\
\hline
Polite--auth. & $2.88$ ($0.84$) & $3.02$ ($0.81$) & $2.96$ ($0.81$) & $3.02$ ($0.83$) \\
\hline
Symp.--anti. & $2.36$ ($0.88$) & $2.22$ ($0.88$) & $2.26$ ($0.87$) & $2.27$ ($0.87$) \\
\hline
\end{tabular}
\end{table}

This descriptive pattern was supported by the mixed-model analysis. When the robot decelerated during the avoidance phase, it was perceived as significantly more hesitant on the \emph{confident--hesitant} dimension ($p<.001$), more fragile on the \emph{solid--fragile} dimension ($p=.029$), weaker on the \emph{strong--weak} dimension ($p=.001$), rougher on the \emph{smooth--rough} dimension ($p=.037$)
. By contrast, interaction distance had no strong main effect on most perceptual dimensions. A non-significant but interpretable tendency was nevertheless observed for \emph{authoritarian--polite} ($p=.080$), suggesting slightly more positive ratings for larger clearances, and a marginal interaction was found for \emph{sympathetic--antipathetic} ($p=.066$), indicating that sympathy may depend on a combination of velocity and distance rather than on either factor alone. These observations are coherent with prior studies showing that robot velocity, smoothness, and interpersonal distance strongly contribute to perceived comfort, safety, and acceptability~\cite{Lasota17,Sisbot2007,Story2018}.

\begin{table}[t]
\caption{Main inferential results for the user study}
\label{tab:main_results}
\centering
\begin{tabular}{|l|l|}
\hline
Variable & Main result \\
\hline
Global NASA-TLX & No effect of velocity, distance, or interaction \\
\hline
Weighted NASA-TLX & No effect of velocity, distance, or interaction \\
\hline
Confident--hesitant & Velocity effect, $p<.001$ \\
\hline
Solid--fragile & Velocity effect, $p=.029$ \\
\hline
Strong--weak & Velocity effect, $p=.001$ \\
\hline
Smooth--rough & Velocity effect, $p=.037$ \\
\hline
Polite--authoritarian & Distance tendency, $p=.080$ \\
\hline
Sympathetic--antipathetic & Velocity $\times$ distance tendency, $p=.066$ \\
\hline
\end{tabular}
\end{table}

\subsection{Interpretation}

Taken together, the two experiments lead to a coherent interpretation. Experiment~1 showed that the proposed control architecture can effectively generate differentiated motion regimes from the pair $(v_{max}, d_{max})$. Experiment~2 then showed that these differences are not only technical: they are also perceived by human users, but mainly through the temporal dynamics of the motion rather than through geometric clearance alone.

The absence of significant workload effects is important. It indicates that none of the tested behaviors made the interaction more mentally or physically demanding in a substantial way. In other words, the proposed behavior modulation does not seem to come at the price of increased subjective difficulty. This is a useful result for robotic design, because it suggests that one can tune the style of the robot behavior without necessarily increasing the human cognitive cost of the interaction.

The perceptual results are arguably even more informative. They show that the velocity profile acts as a strong communicative cue. A robot that accelerates when avoiding the human is interpreted as more confident, more solid, stronger and smoother, and more self-assured. Conversely, a robot that decelerates is more easily interpreted as hesitant, fragile, rough, or doubtful. This means that the temporal profile of the motion does not merely affect kinematic appearance; it changes the social meaning of the robot behavior. This interpretation is in agreement with the literature on legible and expressive motion, which highlights that humans infer robot intent and attitude from dynamic cues such as smoothness, timing, and anticipatory adaptation~\cite{Dragan2013,Hoffman2019}.

By contrast, the weaker effect of interaction distance suggests that geometric clearance alone is not sufficient to strongly alter social perception in the considered task. From the point of view of the controller, this result is particularly interesting. It suggests that the role of $d_{max}$ is primarily to shape anticipation and to maintain conservative safety margins, whereas the role of velocity modulation is more directly linked to legibility and acceptability. In practice, this means that a human-aware controller should not be designed only around separation distances, but also around the way motion unfolds over time. This nuance complements earlier proxemics-oriented studies, where distance was identified as an important factor of comfort and safety, but not necessarily the only or dominant determinant of social interpretation in all task settings~\cite{Sisbot2007,Arai2010,Lasota17}.

Overall, these findings support the main thesis of this paper. Predictive safety-aware motion generation should be evaluated not only in terms of collision avoidance and passive safety guarantees, but also in terms of the subjective impressions it produces during interaction. The proposed parameterization offers a simple and operational way to address both aspects: $d_{max}$ supports conservative and anticipatory avoidance, while velocity modulation influences how the robot is interpreted by the human partner. This makes the proposed framework relevant not only for safe interaction, but also for fluent and acceptable motion generation in shared workspaces, in line with broader perspectives on human-aware navigation and collaborative fluency~\cite{Hoffman2019,Lasota17,Sisbot2007}.

\section{Conclusion and perspectives}
This paper addressed motion generation for robots operating in shared human--robot environments, where safety must be ensured while preserving interaction quality. We proposed an MPC-based framework that combines predictive safety constraints with motion parameters related to human perception, in particular interaction distance and velocity profile.

The experiments showed that the proposed controller can generate distinct robot behaviors and that these behaviors are perceived differently by human users. More specifically, the results suggest that velocity modulation has a stronger impact on perceived social acceptability than geometric clearance alone. This highlights that safe robot motion should be evaluated not only through collision avoidance, but also through the social meaning conveyed by motion timing and smoothness.

These results support the relevance of combining safety, fluency, and acceptability in a unified control framework. 

Future work will focus on a deeper understanding of the impact of these different behaviors, with a particular emphasis on perceived safety and trust, in order to design robot behaviors that can be adaptive and personalized. This will enable the extension of this work to real industrial scenarios involving collaborative robots, where social acceptability, perceived safety, and trust play a key role in long-term human–robot collaboration.

\bibliographystyle{IEEEtran}
\bibliography{roman26}

@ARTICLE{Zheng22, 
author={Zheng, P. and Wieber, P-B. and Baber, J. and Aycard, O.}, 
JOURNAL = {Sensors journal},
SERIES = {Special issue on "Deep Learning-Based Human Intention and Trajectory Prediction Systems Using Sensors"}, 
title={Human arm motion prediction for collision avoidance in a shared workspace}, 
year={2022}, 
doi={10.3390/s22186951}, 
}

@article{Arai2010,
  author={Arai, T. et al.},
  title={Human-Robot Collaborative Systems},
  journal={IEEE Transactions on Industrial Electronics},
  year={2010}
}

@article{Story2018,
  author={Story, M. et al.},
  title={Effects of Robot Motion Characteristics on Human Perception},
  journal={International Journal of Social Robotics},
  year={2018}
}

@article{Howell2019,
  author={Howell, S. et al.},
  title={Workspace overlap and cognitive load in human--robot collaboration},
  journal={IEEE Transactions on Human-Machine Systems},
  year={2019}
}

@INPROCEEDINGS{Svarny2019IROS, 
author={P. {Svarny} and M. {Tesar} and J. K. {Behrens} and M. {Hoffmann}}, 
booktitle={2019 IEEE/RSJ International Conference on Intelligent Robots and Systems (IROS)}, 
title={Safe physical {HRI}: Toward a unified treatment of speed and separation monitoring together with power and force limiting}, 
year={2019}, 
volume={}, 
number={}, 
pages={7580-7587}, 
doi={10.1109/IROS40897.2019.8968463}, 
ISSN={2153-0858}, 
month={Nov},}

@article{albu07,
	title={The DLR lightweight robot: design and control concepts for robots in human environments},
	author={Albu-Sch{\"a}ffer, Alin and Haddadin, Sami and Ott, Ch and Stemmer, Andreas and Wimb{\"o}ck, Thomas and Hirzinger, Gerhard},
	journal={Industrial Robot: an international journal},
	volume={34},
	number={5},
	pages={376--385},
	year={2007},
	publisher={Emerald Group Publishing Limited}
}

@article{26:bouraine2012provably,
	title={Provably safe navigation for mobile robots with limited field-of-views in dynamic environments},
	author={Bouraine, Sara and Fraichard, Thierry and Salhi, Hassen},
	journal={Autonomous Robots},
	volume={32},
	number={3},
	pages={267--283},
	year={2012},
	publisher={Springer}
}

@inproceedings{22:zheng2020online,
  title={Online optimal motion generation with guaranteed safety in shared workspace},
  author={Zheng, Pu and Wieber, Pierre-Brice and Aycard, Olivier},
  booktitle={ICRA},
  year={2020}
}

@article{Li24,
  author    = {Li, Wei and Hu, Yan and Zhou, Yang and Pham, Duc Truong},
  title     = {Safe human--robot collaboration for industrial settings: a survey},
  journal   = {Journal of Intelligent Manufacturing},
  year      = {2024},
  volume    = {35},
  number    = {5},
  pages     = {2235--2261},
  doi       = {10.1007/s10845-023-02058-4}
}

@article{Hoffman2019,
  author  = {Hoffman, Guy},
  title   = {Evaluating Fluency in Human--Robot Collaboration},
  journal = {IEEE Transactions on Human-Machine Systems},
  year    = {2019},
  volume  = {49},
  number  = {3},
  pages   = {209--218},
  doi     = {10.1109/THMS.2019.2904556}
}

@article{Lasota17,
  author  = {Lasota, Przemyslaw A. and Fong, Terrence and Shah, Julie A.},
  title   = {A Survey of Methods for Safe Human--Robot Interaction},
  journal = {Foundations and Trends in Robotics},
  year    = {2017},
  volume  = {5},
  number  = {4},
  pages   = {261--349},
  doi     = {10.1561/2300000052}
}

@article{Sisbot2007,
  author = {Sisbot, Emmanuel A. and Marin-Urias, Luis F. and Alami, Rachid and Simeon, Thierry},
  title = {A Human-Aware Mobile Robot Motion Planner},
  journal = {IEEE Transactions on Robotics},
  year = {2007},
  volume = {23},
  number = {5},
  pages = {874--883}
}

@inproceedings{Dragan2013,
  author = {Anca D. Dragan and Kenton C. T. Lee and Siddhartha S. Srinivasa},
  title = {Legibility and Predictability of Robot Motion},
  booktitle = {Proceedings of the 8th ACM/IEEE International Conference on Human-Robot Interaction (HRI)},
  year = {2013},
  pages = {301--308},
  doi = {10.1109/HRI.2013.6483603}
}

@article{38:cao2019openpose,
  title={OpenPose: realtime multi-person 2D pose estimation using Part Affinity Fields},
  author={Cao, Zhe and Hidalgo, Gines and Simon, Tomas and Wei, Shih-En and Sheikh, Yaser},
  journal={IEEE Transactions on Pattern Analysis and Machine Intelligence},
  volume={43},
  number={1},
  pages={172--186},
  year={2019},
  publisher={IEEE}
}

@article{39:lugaresi2019mediapipe,
  title={MediaPipe: A Framework for Building Perception Pipelines},
  author={Lugaresi, Camillo and Tang, Jiuqiang and Nash, Hadon and McClanahan, Chris and Uboweja, Esha and Hays, Michael and Zhang, Fan and Chang, Chuo-Ling and Yong, Ming Guang and Lee, Juhyun and others},
  journal={arXiv preprint arXiv:1906.08172},
  year={2019}
}

@inproceedings{40:nowak2021point,
  title={Point Clouds With Color: A Simple Open Library for Matching RGB and Depth Pixels from an Uncalibrated Stereo Pair},
  author={Nowak, Jordan and Fraisse, Philippe and Cherubini, Andrea and Daures, Jean-Pierre},
  booktitle={2021 IEEE International Conference on Multisensor Fusion and Integration for Intelligent Systems (MFI)},
  pages={1--7},
  year={2021},
  organization={IEEE}
}

\end{document}